\documentclass{article}
\usepackage{spconf,amsmath,graphicx}
\makeatletter
\def\ninept{%
  \def\baselinestretch{1.05}
  \renewcommand\normalsize{%
    \@setfontsize\normalsize{9.6pt}{11pt}%
  }%
  \normalsize
}
\makeatother

\usepackage{amsmath,amssymb,amsfonts}
\usepackage{algorithmic}
\usepackage{graphicx}
\usepackage{textcomp}
\usepackage{algorithm}
\usepackage{comment}
\usepackage{xcolor}
\usepackage{subcaption} 
\usepackage[table]{xcolor}
\def\BibTeX{{\rm B\kern-.05em{\sc i\kern-.025em b}\kern-.08em
    T\kern-.1667em\lower.7ex\hbox{E}\kern-.125emX}}

\usepackage{booktabs}
\usepackage{multirow}
\usepackage{siunitx} 
\usepackage{pifont} 
\usepackage{fontawesome5} 
\usepackage{subcaption} 
\usepackage{soul} 
\usepackage[hidelinks]{hyperref}
\usepackage[numbers,sort&compress]{natbib}
\usepackage[most]{tcolorbox}

\definecolor{green}{RGB}{241,196,213}

\begin{document}

\title{Conditional Random Fields for Interactive Refinement of Histopathological Predictions
}

\name{\parbox{\textwidth}{\centering
    Tiffanie Godelaine$^\dagger$$^{,1}$\qquad
    Maxime Zanella$^\dagger$$^{,1,2}$\qquad
    Karim El Khoury$^{1}$\\
    {\it Sa\"id Mahmoudi}$^{2}$\qquad
    {\it Beno\^it Macq}$^{1}$\qquad
    {\it Christophe De Vleeschouwer}$^{1}$%
}
}
\address{
    $^{1}$ICTEAM, UCLouvain, Belgium \qquad
    $^{2}$ILIA, UMons, Belgium
}

\ninept
\maketitle

\renewcommand{\thefootnote}{\fnsymbol{footnote}}
\makeatletter
\renewcommand\@makefntext[1]{\noindent\normalfont#1} 
\footnotetext{$^\dagger$denotes equal contribution.\vspace{1mm}\\  T. G. is supported by MedReSyst, funded by the Walloon Region and EU-Wallonie 2021-2027 program. M. Z. is funded by the Walloon Region under grant No. 2010235 (ARIAC by DIGITALWALLONIA4.AI).}
\makeatother

\begin{abstract}
Assisting pathologists in the analysis of histopathological images has high clinical value, as it supports cancer detection and staging. In this context, histology foundation models have recently emerged. Among them, Vision-Language Models (VLMs) provide strong yet imperfect zero-shot predictions. We propose to refine these predictions by adapting Conditional Random Fields (CRFs) to histopathological applications, requiring no additional model training.
We present HistoCRF, a CRF-based framework, with a novel definition of the pairwise potential that promotes label diversity and leverages expert annotations. We consider three experiments: without annotations, with expert annotations, and with iterative human-in-the-loop annotations that progressively correct misclassified patches.
Experiments on five patch-level classification datasets covering different organs and diseases demonstrate average accuracy gains of 16.0\% without annotations and 27.5\% with only 100 annotations, compared to zero-shot predictions. 
Moreover, integrating a human in the loop reaches a further gain of 32.6\% with the same number of annotations. 
The code will be made available on \url{https://github.com/tgodelaine/HistoCRF}. 
\end{abstract}


\label{sec:intro}
\vspace{-1mm}

\section{Introduction}
The detailed views of cells and tissue offered by histological Whole Slide Images (WSIs) are crucial for diagnosing cancers. However, manually annotating theses slides is a labor-intensive process. In particular, slides may contain multiple cancer sites with different grades.
Therefore, automating tissue-type classification is essential to reduce the burden on pathologists. In response, several histology-oriented foundation models have been introduced \cite{Huang2023, modelconch, modeluni}. 
Among them, Vision-Language Models (VLMs), with their strong generalization capabilities, provide patch-level zero-shot predictions on several downstream tasks. However, such predictions are prone to inaccuracies and thus require refinement \cite{Sikaroudi2023}. 
To address this limitation, we propose a method that refines the outputs of VLMs without additional training, and can include real-time guidance from pathologists. 

\begin{figure*}[h]
    \centering
    \includegraphics[width=0.9\linewidth]{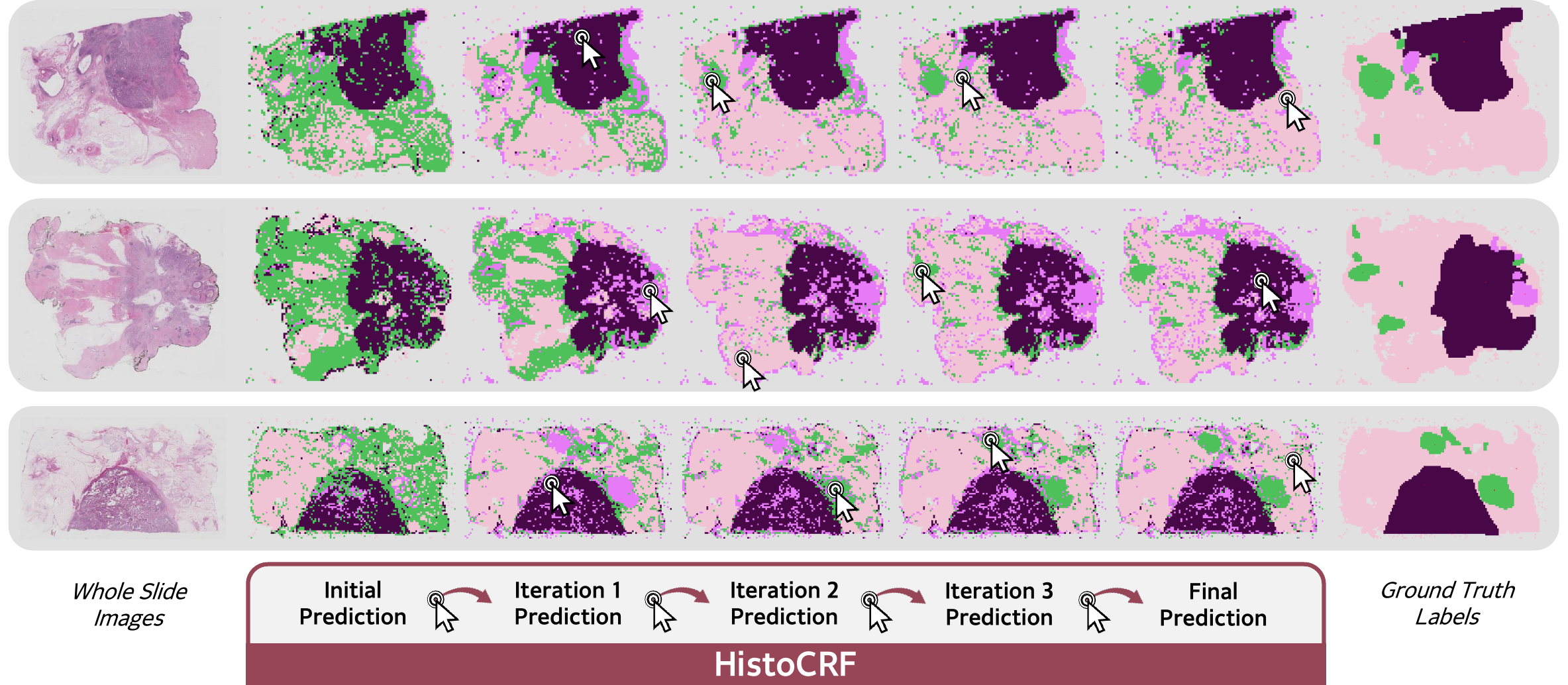}
    \caption{Refinement of histopathological zero-shot predictions on three WSIs of breast cancer tissue \cite{datasetbach} using the proposed method applied to patches extracted from each WSI in the HITL setting. After each prediction, the pathologist annotates a region indicated by the white arrow. These annotations then guide the refinement step leading to improved predictions.}
    \label{fig:visualization}
\end{figure*}

Commonly, in histology, patch-level predictions produced by VLMs are refined by treating each patch independently \cite{Lu2024, Lee2025}.
In contrast, a promising direction is to enhance all predictions at once, leveraging the relationships between patches to provide a better understanding of the WSI \cite{Shao2021, histo_transclip}.
Conditional Random Fields (CRFs) naturally fit into this setting, as they build probabilistic models that capture dependencies between patches, enforcing consistency across predictions while preserving meaningful local differences \cite{ BenAyed2023}. 
CRFs were first used as post-processing for mask smoothing \cite{Krahenbuhl2012, Veksler2025}, before being integrated with deep learning models \cite{Zheng2015}.
In histology, some works \cite{Sun2020, Li2020} train a model and then construct CRFs from its outputs, while others integrate CRFs directly into deep networks and optimize their parameters jointly \cite{Yi2018,Hui2019,ZormpasPetridis2019}. 
In contrast, our approach is directly built on the outputs of histology-oriented VLMs and adapt CRFs to this setting. 
Specifically, we introduce carefully designed pairwise potential terms to promote label diversity and to enable pathologist annotations to guide the refinement process. This proves essential for effective prediction refinement across various annotation settings, including an iterative human-in-the-loop (HITL) scenario. A visual example of our interactive refinement method is shown in Fig. \ref{fig:visualization}.

\textbf{Contributions.} We introduce HistoCRF, a
class prediction refinement procedure, designed to enhance predictions from histology-oriented VLMs. Additionally, we adapt CRFs for histological applications by introducing novel terms, including one that utilizes expert annotations to guide the refinement process. HistoCRF achieves average improvements of 16.0\% without annotations and 27.5\% with 100 annotations across five histopathology benchmark datasets. Its real-time performance highlights its suitability in HITL setting, where average accuracy increases to 32.6\%.

\vspace{-2.5mm}

\label{sec:background}
\section{Background}
\vspace{-1mm}

We define a labeling problem with a set of data instances, associated with a hidden random variable $\mathbf{Y}$ (e.g., the class label) and observed variables $\mathbf{X}$ (e.g., an embedding vector from a foundation model for each data instance). We want to obtain an assignment $\mathbf{y}=(y_v)_{v \in \mathcal{V}}$, meaning we want to infer a class label for each data instance. In a CRF, we use a graph representation $\mathcal{G}=\{\mathcal{V}, \mathcal{E}\}$ where $\mathcal{V}$ is the set of vertices (the data instances) and $\mathcal{E}$ is a set of undirected edges connecting each vertex $v$ to its neighbors $\mathcal{N}_v$ (the relations between these data instances) \cite{BenAyed2023}. \\ 
In CRF, the joint distribution over the hidden random variables is commonly expressed as a Gibbs distribution:
\begin{equation}
    P(\mathbf{Y}=\mathbf{y}) = \frac{1}{Z}\exp(-E(\mathbf{y})), 
\end{equation}
where $Z$ is a normalizing factor and $E(\mathbf{y})=\sum_{c \in \mathcal{C}} \Phi_c(y_c)$ is the energy defined on the set of maximal cliques $\mathcal{C}$ (a clique is a set of fully connected sub-graphs).

If we state our problem as a Bayesian estimation $P(\mathbf{Y}=\mathbf{y}|\mathbf{X}) \propto P(\mathbf{X}|\mathbf{Y}=\mathbf{y})P(\mathbf{Y}=\mathbf{y})$ and taking minus the logarithm, we obtain an objective function of the general form:

\begin{equation}
\label{eq:obj_fun}
\begin{split}
    f(\mathbf{y}) &= - \sum_{v\in\mathcal{V}} \log P(x_v|Y_v=y_v) + E(\mathbf{y})  \\
    &= \underbrace{\sum_{v \in \mathcal{V}} \phi_v (y_v)}_{\substack{\text{unary potentials $\Phi_u$}}} + \underbrace{\sum_{c \in \mathcal{C}} \Phi_c(y_c)}_{\substack{\text{pairwise potentials $\Phi_p$}}}.
\end{split}
\end{equation}
Unary potentials indicate what value each variable is likely to take based on its own observed variables, while pairwise potentials encourage neighboring variables to take compatible values. 
These two potentials are the building blocks of CRF methods, and reflect various design choices \cite{ZormpasPetridis2019,Sun2020}, as we will see in Section \ref{sec:method}. 

As the solution of Eq. \eqref{eq:obj_fun} is intractable, a mean field approximation is used. This minimizes the KL-divergence between 
$P(\mathbf{Y}|\mathbf{X})$ and a simpler approximation $Q$ that can be written as the product of independent marginal distributions  $Q(\mathbf{y)}=\prod_v Q_v(y_v)$.

\begin{algorithm}[h]
\resizebox{0.9\linewidth}{!}{%
\begin{minipage}{\linewidth}
\caption{HistoCRF - HITL}
\label{alg:it}

\textbf{For} $1 \to \text{Number of expert-annotation steps}$

\hspace*{1em} Get pathologist annotations $\mathcal{A}$

\hspace*{1em} Initialize unary potential $\phi^{(0)} \leftarrow \text{Eq.}~\eqref{eq:unary}$

\colorbox{green!30}{%
\parbox{\dimexpr\linewidth-2\fboxsep}{%
    \textcolor{black}{\hspace{0.3cm}\textit{Iterative message passing}}\\
    \hspace*{1em} \textbf{For} $i = 1 \to 50$\\ 
    \hspace*{2em} Get pairwise potential $\Phi_p \leftarrow \text{Eq.}~\eqref{eq:pairwise_potential}$\\
    \hspace*{2em} Iterative update: $Q^{(i)} \leftarrow \text{Eq.}~\eqref{eq:up}$\\
    \hspace*{2em} Update unary potential: $\phi^{(i)} \leftarrow \frac{1}{2}\left(\phi^{(i-1)} + Q^{(i)}\right)$
}}
\textbf{Output:} Predictions $\hat{y} \leftarrow \operatorname*{arg\,max} Q^{(50)}$ \\
\end{minipage}
}
\vspace{-2ex}
\end{algorithm}
This results in an iterative update equation corresponding to a message passing algorithm \cite{Krahenbuhl2012} on the graph  that can be written as: 
\begin{equation}
    \label{eq:up}
    Q_v(y_v=l)=\frac{1}{Z_v}\exp\left(-\phi_v(y_v=l) - \sum_{w \in \mathcal{N}_v}\phi_{vw}(y_v,y_w)\right).
\end{equation}

Hence, assigning label $l$ to vertex $v$ directly depends on its \textit{a priori} unary potential and the compatibility of this assignment with its neighbors $\mathcal{N}_v$.

\label{sec:method}

\vspace{-3mm}
\section{Method}
We propose an adaptation of CRFs for histology image analysis. 
To handle large-scale histopathological images, we divide them into a set of patches. The WSI can then be represented by a graph with each patch being a vertex $v$. Each patch is associated with an unknown class label (hidden random variables $\mathbf{Y}$) and an embedding vector obtained with foundation models (observed variables $\mathbf{X}$). \\
In the following, we describe our formulation of the unary and pairwise potentials and the adaptation of the CRF model for this setting. The key steps of the method are outlined in Algorithm~\ref{alg:it}.

\vspace{-3mm}
\subsection{Unary potentials}
\vspace{-1mm}
In this work, we leverage histology-oriented VLMs to obtain our unary potential $\phi_v(y_v)$ from patch-level predictions without training a task-specific model. More precisely, we encode each patch $v$ with the visual encoder to get an embedding vector $\mathbf{f}_v$ and we obtain for each class $l$ textual embeddings $\mathbf{t}_l$ with the textual encoder from the average of a set of prompts (e.g., \textit{"a histology image of \{class\}"}) \cite{histo_transclip}.
The unary potential is then computed from the cosine similarity:
\vspace{-1mm}
\begin{equation}
\label{eq:unary}
\phi_v(y_v = l)
= - \log \left(
\operatorname{softmax}_l \left(
\frac{\mathbf{f}_v \mathbf{t}_l^\top}{\|\mathbf{f}_v\| \, \|\mathbf{t}_l\|}
\right)
\right).
\end{equation}

\vspace{-7mm}

\subsection{Pairwise potentials}
We decompose the pairwise potential into two complementary terms: (i) a term that encourages label diversity, in contrast to the smoothness term commonly used in the literature, and (ii) a term that incorporates expert-provided annotations, which are typically not exploited in this context.

Both terms are defined based on a sparse neighborhood to avoid quadratic complexity. Neighboring patches, identified using a similarity metric (Eq. \eqref{eq:simi}), are connected under the assumption that they either share the same class or not, depending on the term. 

For the first term, we connect patches that are the most \textit{dissimilar} according to their embedding vectors $\mathbf{f}_v$, extracted using a histology-oriented vision model. By linking patches that appear different but share the same label, this term encourages the refinement process to adjust such inconsistencies, promoting greater label diversity.
This contrasts with the non-local sparse CRF of Veksler and Boykov \cite{Veksler2025}, which connects \textit{similar} pixels based on color intensity.
This leads to the following definition of the first pairwise term: 
\begin{equation}
\label{eq:pairwise_base}
\phi_{vw}(y_v,y_w)=\delta_{y_v,y_w}(1-\text{sim}(\mathbf{f}_v, \mathbf{f}_w)),
\end{equation}
where similarities are measured using the cosine similarity:

\begin{equation}
    \text{sim}(\mathbf{f}_v, \mathbf{f}_w) = \frac{\mathbf{f}_v \mathbf{f}_w^\top}{\|\mathbf{f}_v\|\|\mathbf{f}_w\|}. 
    \label{eq:simi}
\end{equation}

For the second term, we connect annotated patches $\mathcal{A}$ to highly \textit{similar} patches. This encourages the refinement process to align their labels with expert input, promoting consistency with pathologist annotations. 
This leads to the following definition of the second pairwise term: 
\begin{equation}
\label{eq:pairwise_ann}
\psi_{vw}(y_v,y_w)=(1-\delta_{y_v,y_w})\text{sim}(\mathbf{f}_v, \mathbf{f}_w).
\end{equation}

Combining both terms gives our pairwise potential:
\begin{align}
\label{eq:pairwise_potential}
\Phi_p =\underbrace{ \alpha\sum_{v \in \mathcal{V}} \sum_{w \in \mathcal{N}_v} \phi_{vw}(y_v, y_w)}_{\text{(i) base pairwise term}}
+ 
    \underbrace{\beta \sum_{v \in \mathcal{A}} \sum_{w \in \mathcal{M}_v} \psi_{vw}(y_v,y_w)}_{\text{(ii) annotation pairwise term}},
\end{align}
\vspace{-1mm}
where $\alpha$ and $\beta$ are weighting factors. $\mathcal{N}_v$ denotes the set of patches that are among the most \textit{dissimilar} to patch $v \in \mathcal{V}$, while $\mathcal{M}_v$ represents the set of patches that are among the most \textit{similar} to an annotated patch $v \in \mathcal{A}$. 
The neighborhoods $\mathcal{N}_v$ and $\mathcal{M}_v$ are not fixed but reselected each time the pairwise potential is computed. Specifically, they are respectively drawn at random from within a subset of the most dissimilar or most similar patches to $v$. This strategy increases the number of possible connections over iterations without raising memory load.

\label{sec:exp}

\begin{table}[!t]
    \caption{Average accuracy using the \textit{random} sampling strategy applied on five histology classification datasets with varying annotations $|\mathcal{A}|$. ZS denotes the zero-shot accuracy. Experiments are run over 20 seeds.}
    \label{tab:random_sampling}
    \vspace{-1mm}
    \centering
    \footnotesize
    \sisetup{table-format=2.1, detect-weight=true, detect-family=true} 
    \setlength{\tabcolsep}{3pt}

    \begin{tabular}{@{}p{0.88cm} p{2cm} S S S S S S S@{}}

        & & & \multicolumn{6}{c}{\textbf{$|\mathcal{A}|$}} \\
        \cmidrule(l){4-9}
        \textbf{Dataset} & \textbf{Method} & \multicolumn{1}{c}{\textbf{ZS}} & {\textit{0}} & {\textit{5}} & {\textit{10}} & {\textit{25}} & {\textit{50}} & {\textit{100}} \\
                \specialrule{.05em}{.2em} {.2em} 

        \multirow{2}{*}{\textbf{BACH}}
        & LP \cite{Zhou2003} &  & {-} & 53.9 & 65.5 & 74.8 & 80.6 & 87.6 \\
        & \multicolumn{1}{l}{\cellcolor{green!20}HistoCRF} & \multicolumn{1}{c}{\cellcolor{green!20}\multirow{-2}{*}{60.4}}
        & \multicolumn{1}{S}{\cellcolor{green!20} \bfseries 74.4} & \multicolumn{1}{S}{\cellcolor{green!20} \bfseries 78.6} & \multicolumn{1}{S}{\cellcolor{green!20} \bfseries 78.8} & \multicolumn{1}{S}{\cellcolor{green!20} \bfseries 82.9} & \multicolumn{1}{S}{\cellcolor{green!20} \bfseries 85.9} & \multicolumn{1}{S}{\cellcolor{green!20} \bfseries 90.2} \\
        \specialrule{.05em}{.2em} {.2em} 

        \multirow{2}{*}{\textbf{BRACS}}
        & LP \cite{Zhou2003} &  & {-} & 28.7 & 36.2 & 44.5 & 53.0 & 62.5 \\
        & \multicolumn{1}{l}{\cellcolor{green!20}HistoCRF} & \multicolumn{1}{c}{\cellcolor{green!20}\multirow{-2}{*}{31.7}} & \multicolumn{1}{S}{\cellcolor{green!20} \bfseries 47.2} & \multicolumn{1}{S}{\cellcolor{green!20} \bfseries 48.2} & \multicolumn{1}{S}{\cellcolor{green!20} \bfseries 50.4} & \multicolumn{1}{S}{\cellcolor{green!20} \bfseries 52.7} & \multicolumn{1}{S}{\cellcolor{green!20} \bfseries 57.0} & \multicolumn{1}{S}{\cellcolor{green!20} \bfseries 63.5} \\
        \specialrule{.05em}{.2em} {.2em} 

        \multirow{2}{*}{\textbf{ESCA}}
        & LP \cite{Zhou2003} &  & {-} & {-} & {-} & {-} & {-} & {-} \\
        & \multicolumn{1}{l}{\cellcolor{green!20}HistoCRF} & \multicolumn{1}{c}{\cellcolor{green!20}\multirow{-2}{*}{63.3}} & \multicolumn{1}{S}{\cellcolor{green!20} \bfseries 67.4} & \multicolumn{1}{S}{\cellcolor{green!20} \bfseries 67.5} & \multicolumn{1}{S}{\cellcolor{green!20} \bfseries 67.5} & \multicolumn{1}{S}{\cellcolor{green!20} \bfseries 67.5} & \multicolumn{1}{S}{\cellcolor{green!20} \bfseries 67.7} & \multicolumn{1}{S}{\cellcolor{green!20} \bfseries 67.9} \\
        \specialrule{.05em}{.2em} {.2em} 

        \multirow{2}{*}{\textbf{NCT}}
        & LP \cite{Zhou2003} &  & {-} & 53.9 & 68.8 & \bfseries 85.5 & \bfseries 92.9 & \bfseries 96.0 \\
        & \multicolumn{1}{l}{\cellcolor{green!20}HistoCRF} & \multicolumn{1}{c}{\cellcolor{green!20}\multirow{-2}{*}{60.5}} & \multicolumn{1}{S}{\cellcolor{green!20} \bfseries 72.1} & \multicolumn{1}{S}{\cellcolor{green!20} \bfseries 74.0} & \multicolumn{1}{S}{\cellcolor{green!20} \bfseries 75.3} & \multicolumn{1}{S}{\cellcolor{green!20} 78.0} & \multicolumn{1}{S}{\cellcolor{green!20} 81.1} & \multicolumn{1}{S}{\cellcolor{green!20} 86.9} \\
        \specialrule{.05em}{.2em} {.2em} 

        \multirow{2}{*}{\textbf{SICAP}}
        & LP \cite{Zhou2003} &  & {-} & 50.0 & 54.3 & 61.4 & 63.4 & 68.1 \\
        & \multicolumn{1}{l}{\cellcolor{green!20}HistoCRF} & \multicolumn{1}{c}{\cellcolor{green!20}\multirow{-2}{*}{27.4}} & \multicolumn{1}{S}{\cellcolor{green!20} \bfseries 61.8} & \multicolumn{1}{S}{\cellcolor{green!20} \bfseries 62.3} & \multicolumn{1}{S}{\cellcolor{green!20} \bfseries 62.6} & \multicolumn{1}{S}{\cellcolor{green!20} \bfseries 64.1} & \multicolumn{1}{S}{\cellcolor{green!20} \bfseries 65.4} & \multicolumn{1}{S}{\cellcolor{green!20} \bfseries 68.9} \\
        \specialrule{.2em}{.2em} {.2em} 
        
        \multirow{2}{*}{\textbf{AVG.}}
        & LP \cite{Zhou2003} &  & {-} & 46.6 & 56.2 & 66.5 & \bfseries 72.5 & \bfseries 78.6 \\
        & \multicolumn{1}{l}{\cellcolor{green!20}HistoCRF} & \multicolumn{1}{c}{\cellcolor{green!20}\multirow{-2}{*}{48.6}} & \multicolumn{1}{S}{\cellcolor{green!20} \bfseries 64.6} & \multicolumn{1}{S}{\cellcolor{green!20} \bfseries 66.7} & \multicolumn{1}{S}{\cellcolor{green!20} \bfseries 66.9} & \multicolumn{1}{S}{\cellcolor{green!20} \bfseries 69.1} & \multicolumn{1}{S}{\cellcolor{green!20} 71.4} & \multicolumn{1}{S}{\cellcolor{green!20} 75.5} \\

    \end{tabular}%
\end{table}

\begin{table}[!t]
    \caption{Average accuracy using the \textit{error-based} sampling strategy applied on five histology classification datasets. In the HITL setting, the expert annotates in steps of 5 patches up to 100.}
    \label{tab:hitl_sampling}
    \vspace{-1mm}
    \centering
    \footnotesize
    \sisetup{table-format=2.1, detect-weight=true, detect-family=true} 
    \setlength{\tabcolsep}{3pt}

    \begin{tabular}{@{}p{0.88cm} p{2cm} S S S S S S S@{}}

        & & & \multicolumn{6}{c}{\textbf{$|\mathcal{A}|$}} \\
        \cmidrule(l){4-9}
        \textbf{Dataset} & \textbf{Method} & \multicolumn{1}{c}{\textbf{ZS}} & {\textit{0}} & {\textit{5}} & {\textit{10}} & {\textit{25}} & {\textit{50}} & {\textit{100}} \\
                \specialrule{.05em}{.2em} {.2em} 

        \multirow{3}{*}{\textbf{BACH}}
        & LP \cite{Zhou2003} &  & {-} & 49.8 & 55.5 & 63.8 & 70.6 & 72.6 \\
        & \multicolumn{1}{l}{\cellcolor{green!20}HistoCRF} &
        \multicolumn{1}{c}{\cellcolor{green!20}} & \multicolumn{1}{S}{\cellcolor{green!20} 74.4} & \multicolumn{1}{S}{\cellcolor{green!20} 80.3} & \multicolumn{1}{S}{\cellcolor{green!20} 81.7} & \multicolumn{1}{S}{\cellcolor{green!20} 84.7} & \multicolumn{1}{S}{\cellcolor{green!20} 86.4} & \multicolumn{1}{S}{\cellcolor{green!20} 87.2} \\
        & \multicolumn{1}{l}{\cellcolor{green!40}HistoCRF + HITL} & \multicolumn{1}{c}{\cellcolor{green!40}\multirow{-3}{*}{60.4}} & \multicolumn{1}{S}{\cellcolor{green!40} \bfseries 74.4} & \multicolumn{1}{S}{\cellcolor{green!40} \bfseries 80.5} & \multicolumn{1}{S}{\cellcolor{green!40} \bfseries 83.1} & \multicolumn{1}{S}{\cellcolor{green!40} \bfseries 88.5} & \multicolumn{1}{S}{\cellcolor{green!40} \bfseries 94.7} & \multicolumn{1}{S}{\cellcolor{green!40} \bfseries 99.6} \\
                \specialrule{.05em}{.2em} {.2em} 

        \multirow{3}{*}{\textbf{BRACS}}
        & LP \cite{Zhou2003} &  & {-} & 25.2 & 30.6 & 38.4 & 46.2 & 56.0 \\
        & \multicolumn{1}{l}{\cellcolor{green!20}HistoCRF} & \multicolumn{1}{c}{\cellcolor{green!20}} & \multicolumn{1}{S}{\cellcolor{green!20} 47.2} & \multicolumn{1}{S}{\cellcolor{green!20} 48.1} & \multicolumn{1}{S}{\cellcolor{green!20} 49.4} & \multicolumn{1}{S}{\cellcolor{green!20} 52.8} & \multicolumn{1}{S}{\cellcolor{green!20} 56.9} & \multicolumn{1}{S}{\cellcolor{green!20} 63.6} \\
        & \multicolumn{1}{l}{\cellcolor{green!40}HistoCRF + HITL} & \multicolumn{1}{c}{\cellcolor{green!40}\multirow{-3}{*}{31.7}} & \multicolumn{1}{S}{\cellcolor{green!40} \bfseries 47.2} & \multicolumn{1}{S}{\cellcolor{green!40} \bfseries 48.1} & \multicolumn{1}{S}{\cellcolor{green!40} \bfseries 49.8} & \multicolumn{1}{S}{\cellcolor{green!40} \bfseries 55.4} & \multicolumn{1}{S}{\cellcolor{green!40} \bfseries 61.1} & \multicolumn{1}{S}{\cellcolor{green!40} \bfseries 69.2} \\
                \specialrule{.05em}{.2em} {.2em} 

        \multirow{3}{*}{\textbf{ESCA}}
        & LP \cite{Zhou2003} &  & {-} & {-} & {-} & {-} & {-} & {-} \\
        & \multicolumn{1}{l}{\cellcolor{green!20}HistoCRF} & \multicolumn{1}{c}{\cellcolor{green!20}} & \multicolumn{1}{S}{\cellcolor{green!20} 67.4} & \multicolumn{1}{S}{\cellcolor{green!20} 67.5} & \multicolumn{1}{S}{\cellcolor{green!20} 67.5} & \multicolumn{1}{S}{\cellcolor{green!20} 67.6} & \multicolumn{1}{S}{\cellcolor{green!20} 67.8} & \multicolumn{1}{S}{\cellcolor{green!20} 68.2} \\
        & \multicolumn{1}{l}{\cellcolor{green!40}HistoCRF + HITL} & \multicolumn{1}{c}{\cellcolor{green!40}\multirow{-3}{*}{63.3}} & \multicolumn{1}{S}{\cellcolor{green!40} \bfseries 67.4} & \multicolumn{1}{S}{\cellcolor{green!40} \bfseries 67.5} & \multicolumn{1}{S}{\cellcolor{green!40} \bfseries 67.6} & \multicolumn{1}{S}{\cellcolor{green!40} \bfseries 67.8} & \multicolumn{1}{S}{\cellcolor{green!40} \bfseries 68.0} & \multicolumn{1}{S}{\cellcolor{green!40} \bfseries 68.4} \\
                \specialrule{.05em}{.2em} {.2em} 

        \multirow{3}{*}{\textbf{NCT}}
        & LP \cite{Zhou2003} &  & {-} & 49.2 & 64.2 & 72.2 & 78.0 & 82.6 \\
        & \multicolumn{1}{l}{\cellcolor{green!20}HistoCRF} & \multicolumn{1}{c}{\cellcolor{green!20}} & \multicolumn{1}{S}{\cellcolor{green!20} 72.1} & \multicolumn{1}{S}{\cellcolor{green!20} 74.5} & \multicolumn{1}{S}{\cellcolor{green!20} \bfseries 76.7} & \multicolumn{1}{S}{\cellcolor{green!20} 80.6} & \multicolumn{1}{S}{\cellcolor{green!20} 87.3} & \multicolumn{1}{S}{\cellcolor{green!20} 95.0} \\
        & \multicolumn{1}{l}{\cellcolor{green!40}HistoCRF + HITL} & \multicolumn{1}{c}{\cellcolor{green!40}\multirow{-3}{*}{60.5}} & \multicolumn{1}{S}{\cellcolor{green!40} \bfseries 72.1} & \multicolumn{1}{S}{\cellcolor{green!40} \bfseries 74.5} & \multicolumn{1}{S}{\cellcolor{green!40} 76.3} & \multicolumn{1}{S}{\cellcolor{green!40}\bfseries  80.8} & \multicolumn{1}{S}{\cellcolor{green!40} \bfseries 92.9} & \multicolumn{1}{S}{\cellcolor{green!40} \bfseries 96.9} \\
                \specialrule{.05em}{.2em} {.2em} 

        \multirow{3}{*}{\textbf{SICAP}}
        & LP \cite{Zhou2003} &  & {-} & 49.3 & 55.8 & 60.8 & 62.9 & 65.4 \\
        & \multicolumn{1}{l}{\cellcolor{green!20}HistoCRF} & \multicolumn{1}{c}{\cellcolor{green!20}} & \multicolumn{1}{S}{\cellcolor{green!20} 61.8} & \multicolumn{1}{S}{\cellcolor{green!20} 63.2} & \multicolumn{1}{S}{\cellcolor{green!20} 63.8} & \multicolumn{1}{S}{\cellcolor{green!20} 65.0} & \multicolumn{1}{S}{\cellcolor{green!20} 65.2} & \multicolumn{1}{S}{\cellcolor{green!20} 66.4} \\
        & \multicolumn{1}{l}{\cellcolor{green!40}HistoCRF + HITL} & \multicolumn{1}{c}{\cellcolor{green!40}\multirow{-3}{*}{27.4}} & \multicolumn{1}{S}{\cellcolor{green!40} \bfseries 61.8} & \multicolumn{1}{S}{\cellcolor{green!40} \bfseries 63.2} & \multicolumn{1}{S}{\cellcolor{green!40} \bfseries 64.1} & \multicolumn{1}{S}{\cellcolor{green!40} \bfseries 64.7} & \multicolumn{1}{S}{\cellcolor{green!40} \bfseries 68.0} & \multicolumn{1}{S}{\cellcolor{green!40} \bfseries 72.1} \\
        \specialrule{.2em}{.2em} {.2em} 
        
        \multirow{3}{*}{\textbf{AVG.}}
        & LP \cite{Zhou2003} &  & {-} & 43.4 & 51.5 & 58.8 & 64.4 & 69.2 \\
        & \multicolumn{1}{l}{\cellcolor{green!20}HistoCRF} & \multicolumn{1}{c}{\cellcolor{green!20}} & \multicolumn{1}{S}{\cellcolor{green!20} 64.6} & \multicolumn{1}{S}{\cellcolor{green!20} 66.8} & \multicolumn{1}{S}{\cellcolor{green!20} 67.8} & \multicolumn{1}{S}{\cellcolor{green!20} 70.1} & \multicolumn{1}{S}{\cellcolor{green!20} 72.7} & \multicolumn{1}{S}{\cellcolor{green!20} 76.1} \\
        & \multicolumn{1}{l}{\cellcolor{green!40}HistoCRF + HITL} & \multicolumn{1}{c}{\cellcolor{green!40}\multirow{-3}{*}{48.6}} & \multicolumn{1}{S}{\cellcolor{green!40} \bfseries 64.6} & \multicolumn{1}{S}{\cellcolor{green!40} \bfseries 66.8} & \multicolumn{1}{S}{\cellcolor{green!40} \bfseries 68.2} & \multicolumn{1}{S}{\cellcolor{green!40} \bfseries 71.4} & \multicolumn{1}{S}{\cellcolor{green!40} \bfseries 76.9} & \multicolumn{1}{S}{\cellcolor{green!40} \bfseries 81.2} \\

    \end{tabular}%
\end{table}

\vspace{-2mm}
\section{Experiments}
\vspace{-1mm}
\textbf{Datasets.} Our method is tested on five histology classification datasets covering diverse organs. They contain patches of varying sizes (from 224$\times$224 to 1536$\times$2048) extracted from several WSIs. 
BACH \cite{datasetbach} dataset contains 400 patches from four different types of breast tumour. BRACS \cite{datasetbracs} dataset is composed of 568 patches of seven types of breast lesions. ESCA \cite{datasetesca} dataset contains $\sim10^5$ patches of 11 tissue types of oesophageal adenocarcinoma. NCT \cite{datasetnct} dataset contains $\sim7\cdot10^3$ patches from nine cancer and normal tissues. SICAP \cite{datasetsicapmil} dataset consists of 1500 patches of prostate cancers labels associated to five Gleason score. 

\noindent \textbf{Implementation details. } 
The VLM considered to define the unary potential is CONCH \cite{modelconch}. We use an other model, the vision encoder UNI-2h \cite{modeluni}, to extract stronger features to define similarities between patches in the pairwise potential.
The weighting factors are set as $\alpha=0.1$ and $\beta=0.01$. The base pairwise potential is sparsified to 16 connections per patch ($|\mathcal{N}_v|=16$), and the annotation pairwise potential to 5 ($|\mathcal{M}_v|=5$). Our choices are discussed in Table \ref{tab:ablations}.

\noindent \textbf{Baselines.} We compare HistoCRF with the VLM zero-shot prediction and the label propagation method (LP) introduced by Zhou \textit{et al. }\cite{Zhou2003}, which propagates labels from annotated samples by computing the inverse of a graph Laplacian matrix. For LP, we fix $\alpha$ to 0.5.

\vspace{-3mm}

\subsection{Experiment 1: Without annotations (\texorpdfstring{$|\mathcal{A}|= 0$}{|A|=1})}
The objective is to determine the gain without annotations, where refinement relies solely on the diversity term. 

The average accuracy of HistoCRF and the baselines is shown in Table \ref{tab:random_sampling}. A key advantage of HistoCRF is that it does not require any annotations, yet achieves an average accuracy improvement of 16.0\% compared to the zero-shot predictions.
\vspace{-2mm}

\subsection{Experiment 2: With expert annotations (\texorpdfstring{$|\mathcal{A}|\geq 1$}{|A|≥1})}
The objective is to leverage expert-provided annotations that are incorporated through the annotation pairwise term ($\beta=0.01$). We study two annotation sampling strategies: \textit{random} and \textit{error-based}. In \textit{random}, patches are annotated at random whereas in \textit{error-based}, a set of misclassified patches is annotated.

The average accuracy using \textit{random} and \textit{error-based} annotation sampling strategies are summarized in Tables \ref{tab:random_sampling} and \ref{tab:hitl_sampling}, respectively. HistoCRF outperforms LP in most cases, particularly when the number of annotations is low. 
Comparing both annotation sampling strategies, HistoCRF with \textit{error-based} leads to better results while LP shows reduced average accuracies. 
Note that LP consumes more memory due to the inverse operation, causing failure to produce results for ESCA dataset.  
\vspace{-2mm}

\subsection{Experiment 3: Human-in-the-loop}
The objective is to determine the value of integrating a human in the loop. In HistoCRF-HITL, a pathologist is simulated via an oracle that annotates 5 misclassified patches at the start and after each message passing iteration, until 100 patches are annotated.

Results are presented in Table \ref{tab:hitl_sampling}. Iterative correction provides more effective guidance for refinement, yielding an average accuracy gain from 0.4\% with 10 annotations to up to 5.1\% with 100 annotations, compared to default setting. 
Interestingly, as shown in Table~\ref{tab:time_transposed}, one iterative message passing for a WSI as in Fig. \ref{fig:visualization} (of about $10^4$ patches) takes 2.5 seconds, enabling real-time integration of pathologist annotations.

\vspace{-3mm}
\subsection{Ablation studies: Impact of pairwise potential terms}

We evaluate the contribution of each term in the pairwise potential to prediction accuracy. The results are summarized in Table \ref{tab:ablations}. We observe that the diversity-enforcing term (Eq. \eqref{eq:pairwise_base}) generally outperforms the conventional smoothing term (Eq. \eqref{eq:pairwise_ann}), particularly when the number of annotations is low. 
Decreasing sparsity generally improves accuracy but comes at the cost of higher memory consumption. We identify a trade-off that enables real-time performance while maintaining high performance. 
Finally, we find that incorporating the annotation pairwise term further improves accuracy.

\label{sec:conclu}
\begin{table}[!t]
\centering
\footnotesize
\caption{HistoCRF runtime on a NVIDIA A10 GPU for one iterative message passing.}
\vspace{-2mm}
\label{tab:time_transposed}
\begin{tabular}{lccc}
\textbf{} & \textbf{SICAP} & \textbf{NCT} & \textbf{ESCA} \\
\hline
\textbf{Total patches} & $\sim 10^3$ & $\sim 10^4$ & $\sim 10^5$ \\
\textbf{Runtime} & $\sim 1.5$s & $\sim 2.5$s & $\sim 8$s \\
\bottomrule
\end{tabular}
\end{table}
\vspace{-4mm}

\begin{table}[!t]
    \centering
    \footnotesize
    \caption{Ablation study on the terms composing the pairwise potential term in HistoCRF on the BRACS dataset.}
       \vspace{-2mm}

    \label{tab:ablations}

    \setlength{\tabcolsep}{10pt} 


    \begin{subtable}{\linewidth}
        \centering
        \footnotesize
        \caption{Impact of pairwise potential definition $\phi_{vw}$}
        \vspace{-1mm}
        \label{tab:ablations_pp}
        \begin{tabular}{lcccc}
            \multirow{2}{*}{\textbf{$\phi_{vw}$}} & \multicolumn{4}{c}{\textbf{$|\mathcal{A}|$}} \\
            \cmidrule(lr){2-5}
             & \textit{0} & \textit{10} & \textit{50} & \textit{100} \\
            \midrule
            \cellcolor{green!40}Eq. \eqref{eq:pairwise_base} & \cellcolor{green!40}47.2 & \cellcolor{green!40}49.4 & \cellcolor{green!40}56.9 & \cellcolor{green!40}63.6 \\
            Eq. \eqref{eq:pairwise_ann} & 30.3 & 39.1 & 56.0 & 63.4 \\
            \bottomrule
        \end{tabular}
    \end{subtable}

    \vspace{0.5em} 

    \begin{subtable}{\linewidth}
        \centering
        \footnotesize
        \caption{Impact of $\alpha$}
        
        \vspace{-1mm}
        \label{tab:ablations_omegabase}
        \begin{tabular}{lccccc}
           \textbf{$\alpha$} & \textbf{0.01} & \cellcolor{green!40}\textbf{0.1} & \textbf{0.5} & \textbf{1.0} \\
            \midrule
            \textbf{$|\mathcal{A}|=0$}   & 40.0 & \cellcolor{green!40}47.2 & 42.3 & 32.4 \\
            \bottomrule
        \end{tabular}
    \end{subtable}

    \vspace{0.5em} 

    \begin{subtable}{\linewidth}
        \centering
        \footnotesize
        \caption{Impact of $\beta$}
                \vspace{-1mm}
        \label{tab:ablations_omegaann}
        \begin{tabular}{lccccc}
           \textbf{$\beta$} & \textbf{0.0} & \textbf{0.001} & \cellcolor{green!40}\textbf{0.01} & \textbf{0.05} & \textbf{0.1} \\
            \midrule
            \textbf{$|\mathcal{A}|=50$}  & 53.9 & 54.8 & \cellcolor{green!40}56.9 & 55.0 & 53.9 \\
            \textbf{$|\mathcal{A}|=100$} & 59.7 & 60.7 & \cellcolor{green!40}63.6 & 61.1 & 59.9 \\
            \bottomrule
        \end{tabular}
    \end{subtable}

    \vspace{0.5em} 
    
    \begin{subtable}{\linewidth}
        \centering
        \footnotesize
        \caption{Impact of $|\mathcal{N}_v|$}
        \label{tab:ablations_sparsity}
                \vspace{-1mm}
        \begin{tabular}{lccccc}
            \multirow{2}{*}{\textbf{$|\mathcal{N}_v|$}} & \multicolumn{4}{c}{\textbf{$|\mathcal{A}|$}} & \multirow{2}{*}{\textbf{Mem.}} \\
            \cmidrule(lr){2-5}
            & \textit{0} & \textit{10} & \textit{50} & \textit{100} & \\
            \midrule
            \textbf{4}   & 47.2 & 48.0 & 55.9 & 62.8 & $\text{72 kB}$ \\
            \textbf{8}   & 47.2 & 48.8 & 56.7 & 63.5 & $\text{150 kB}$ \\
            \cellcolor{green!40}\textbf{16}  & \cellcolor{green!40}47.2 & \cellcolor{green!40}49.4 & \cellcolor{green!40}56.9 & \cellcolor{green!40}63.6 & \cellcolor{green!40}$\text{250 kB}$ \\
            \textbf{50}  & 47.7 & 49.1 & 57.0 & 64.2 & $\text{750 kB}$ \\
            \textbf{100} & 47.0 & 49.2 & 57.2 & 64.0 & $\text{1.5 MB}$ \\
            \bottomrule
        \end{tabular}
    \end{subtable}

    \vspace{-0.5mm}
\end{table}

\vfill
\section{Conclusion}
\vspace{-2mm}

HistoCRF leverages zero-shot predictions of VLMs and adapts CRFs to the histological field, enabling their real-time refinement. The pairwise term is augmented with novel components that promote label diversity and incorporate pathologist annotations, allowing experts to guide the correction of patch-level predictions and achieve a faster, more accurate understanding of WSIs. While effective at the patch-level, HistoCRF can be extended by adding spatial pairwise terms for WSIs and by moving from patch- to pixel-level classification.



\newpage

\bibliographystyle{IEEEbib}
\bibliography{strings,refs}

\end{document}